\pdfoutput=1
\documentclass[conference]{IEEEtran}
\IEEEoverridecommandlockouts
\usepackage{cite}
\usepackage{amsmath,amssymb,amsfonts}
\usepackage{algorithmic}
\usepackage{graphicx}
\usepackage{textcomp}
\usepackage{xcolor}
\usepackage{float}
\usepackage{booktabs}
\usepackage{tabularx}
\usepackage[hidelinks]{hyperref}
\def\BibTeX{{\rm B\kern-.05em{\sc i\kern-.025em b}\kern-.08em
    T\kern-.1667em\lower.7ex\hbox{E}\kern-.125emX}}
\begin{document}

\title{RAFM-SER++: A Lightweight Multimodal Emotion Recognition Framework for Real-Time Behavioral Monitoring in Surveillance Systems
}

\author{\IEEEauthorblockN{Ngo Truong Dinh\textsuperscript{*}}
\IEEEauthorblockA{\textit{Department of Data Science} \\
\textit{Industrial University of HCM City}\\
Ho Chi Minh City, Vietnam \\
ngodinh2339@gmail.com}
\and
\IEEEauthorblockN{Tung-Lam Bui\textsuperscript{*}}
\IEEEauthorblockA{
\textit{Le Hong Phong High School for the Gifted}\\
Ho Chi Minh City, Vietnam \\
buitunglam2009@gmail.com}
\and
\IEEEauthorblockN{Chi-Trung Duong\textsuperscript{*}}
\IEEEauthorblockA{\textit{Department of Data Science} \\
\textit{Industrial University of HCM City}\\
Ho Chi Minh City, Vietnam \\
duongchitrung1104@gmail.com}
\and
\IEEEauthorblockN{Vien Nguyen Thi}
\IEEEauthorblockA{\textit{Department of Data Science} \\
\textit{Industrial University of HCM City}\\
Ho Chi Minh City, Vietnam \\
vi12091994@gmail.com}
\and
\IEEEauthorblockN{Viet-Anh Nguyen}
\IEEEauthorblockA{\textit{Faculty of Information Technology} \\
\textit{FPT University}\\
Ho Chi Minh City, Vietnam \\
vietanhnguyen742001@gmail.com}
\and
\IEEEauthorblockN{Phuc-Lu Le}
\IEEEauthorblockA{\textit{Faculty of Information Technology} \\
\textit{University of Science, VNU-HCM}\\
Ho Chi Minh City, Vietnam \\
ORCID: 0009-0007-0254-0202}
}

\maketitle
\begingroup
\renewcommand{\thefootnote}{*}
\begin{NoHyper}
\footnotetext{These authors contributed equally to this work.}
\end{NoHyper}
\endgroup
\begin{abstract}
Recent multimodal Speech Emotion Recognition (SER) systems achieve high accuracy through interaction-heavy cross-modal transformers, but their computational cost limits deployment in latency-sensitive and resource-constrained surveillance systems. To address this challenge, we propose \textbf{RAFM\_SER++}, a lightweight multimodal SER framework featuring an \textbf{asymmetric Residual Attention Fusion Mechanism (RAFM)}. Rather than relying on computationally expensive bidirectional interactions, RAFM injects affective speech cues into semantic text representations through a one-directional residual attention pathway. Combined with a BYOL-inspired cross-modal alignment objective and attention-guided pooling, the proposed framework improves multimodal representation learning while maintaining low computational overhead.

Experiments on the IEMOCAP and ESD benchmarks demonstrate that RAFM\_SER++ consistently outperforms the HuBERT-Base baseline and achieves a superior accuracy--efficiency trade-off compared with the state-of-the-art MemoCMT. Specifically, RAFM\_SER++ reduces trainable parameters by more than 60\%, achieves faster inference (79.60 it/s), and attains BACC scores of \textbf{81.10\%} on \textbf{IEMOCAP} and \textbf{95.39\%} on \textbf{ESD}. These results indicate that lightweight asymmetric multimodal fusion is an effective alternative to interaction-heavy cross-modal transformers for real-time surveillance applications.
\end{abstract}


\section{INTRODUCTION}

Audio-visual surveillance systems are widely deployed in public safety, smart cities, transportation hubs, and human-centered monitoring. However, vision-based perception is often affected by occlusion, poor illumination, adverse viewpoints, and privacy constraints, making \textit{audio understanding} an important complementary source of behavioral information.

Speech Emotion Recognition (SER) infers human affective states from vocal characteristics such as pitch, energy, rhythm, and spectral patterns~\cite{el2011speech}. Compared with facial expressions, speech often reflects more spontaneous emotional responses and remains available when visual information is unreliable~\cite{eyben2016geneva}. Consequently, SER has become an important component for surveillance-oriented applications including distress detection, aggression monitoring, conflict escalation awareness, and abnormal behavior analysis.

Recent transformer-based architectures and self-supervised speech encoders such as wav2vec~2.0 and HuBERT have significantly improved SER performance~\cite{baevski2020wav2vec,hsu2021hubert}. Multimodal approaches further enhance robustness by combining acoustic and textual information~\cite{siriwardhana2020jointly,luo2023cross,memoCMT}. However, most existing methods rely on dense bidirectional cross-modal transformers or repeated attention blocks, resulting in high computational cost and inference latency that limit deployment on resource-constrained surveillance platforms.

To address this limitation, we propose \textbf{RAFM\_SER++}, a lightweight multimodal SER framework for real-time behavioral monitoring. Its core component, the \textbf{Residual Attention Fusion Mechanism (RAFM)}, performs asymmetric one-directional residual attention that injects affective speech cues into semantic text representations. Combined with a BYOL-inspired cross-modal alignment objective and attention-guided pooling, RAFM\_SER++ improves multimodal representation learning while avoiding heavy bidirectional interaction.

Experiments on the IEMOCAP and ESD benchmarks show that RAFM\_SER++ achieves competitive or superior recognition performance while reducing trainable parameters by more than 60\% and providing faster inference than transformer-based baselines, demonstrating an effective balance between recognition accuracy and deployment efficiency.

The main contributions of this work are summarized as follows:

\begin{itemize}

\item \textbf{Deployment-oriented multimodal SER.}
We formulate multimodal SER as a lightweight audio intelligence module for behavioral monitoring with emphasis on computational efficiency and real-time deployment.

\item \textbf{Asymmetric Residual Attention Fusion.}
We propose RAFM, a one-directional residual attention mechanism that efficiently injects affective speech information into semantic text representations while reducing redundant bidirectional interaction.

\item \textbf{Lightweight multimodal representation learning.}
We integrate BYOL-inspired cross-modal alignment with attention-guided pooling to improve representation consistency and adaptive feature aggregation.

\item \textbf{Comprehensive evaluation.}
Experiments on IEMOCAP and ESD demonstrate a favorable accuracy--efficiency trade-off, achieving over 60\% fewer trainable parameters and faster inference than stronger transformer-based baselines.

\end{itemize}

\section{RELATED WORK}

Speech Emotion Recognition (SER) is widely used in human-centered intelligent systems, including human--computer interaction, public safety, and behavioral monitoring. Despite substantial progress, achieving both high accuracy and low inference latency remains challenging in noisy, multilingual, and real-world environments~\cite{latif2021survey,akcay2020speech}. Self-supervised speech encoders such as wav2vec~2.0 and HuBERT have improved acoustic representation learning~\cite{baevski2020wav2vec,hsu2021hubert}, while multimodal learning enhances robustness by combining complementary cues from speech, text, and vision~\cite{poria2017review,tsai2019multimodal,zadeh2018cmumosei}. However, heterogeneous modalities differ in semantic abstraction, temporal alignment, and feature distributions, making efficient multimodal fusion a key challenge~\cite{baltrusaitis2019multimodal}.

Multimodal fusion is commonly categorized into early fusion, late fusion, and model-level fusion~\cite{atrey2010multimodal,liu2018efficient}. Early fusion retains rich information but may suffer from high dimensionality, modality imbalance, and overfitting, whereas late fusion is efficient but weak at modeling fine-grained cross-modal dependencies. Therefore, model-level methods based on memory networks and cross-modal attention have become dominant for multimodal representation learning~\cite{zadeh2018memory,ghosal2018contextual,tsai2019multimodal}. Nevertheless, their increasingly dense cross-modal interactions introduce higher computational cost, memory usage, and inference latency, which limits real-time surveillance deployment~\cite{luo2023cross,memoCMT}.

Recent efficiency-oriented studies adopt lightweight language models such as MobileBERT and DistilBERT, together with self-supervised speech encoders including wav2vec~2.0 and HuBERT~\cite{sun2020mobilebert,sanh2019distilbert,baevski2020wav2vec,hsu2021hubert}. Benchmarks and toolkits such as SUPERB/S3PRL further support pretrained speech representations for downstream SER tasks~\cite{yang2021superb}. However, although encoder complexity has been reduced, many multimodal SER systems still rely on interaction-heavy fusion modules such as cross-modal transformers or repeated attention blocks~\cite{luo2023cross,memoCMT}, making the fusion stage a major bottleneck for real-time deployment.

Representative speech--text SER methods include CM-RoBERTa~\cite{luo2023cross}, which uses parallel self-attention and cross-attention to model intra-modal and inter-modal dependencies, and MemoCMT~\cite{memoCMT}, which employs bidirectional Cross-Modal Transformers with memory tokens for richer contextual interaction. While effective, these architectures increase cross-modal interaction capacity at the cost of higher computation, memory consumption, and latency, making deployment on resource-constrained surveillance platforms difficult.

Unlike these interaction-oriented methods, \textbf{RAFM\_SER++} does not aim to maximize cross-modal interaction complexity. Instead, it optimizes the trade-off between recognition performance and deployment efficiency through an asymmetric Residual Attention Fusion Mechanism that injects affective speech cues into semantic text representations via a lightweight one-directional residual attention pathway. Combined with cross-modal alignment and attention-guided pooling, RAFM\_SER++ shows that carefully constrained multimodal interaction can preserve speech--text complementarity while substantially reducing computational overhead for real-time surveillance applications.

\section{METHOD}

\subsection{Framework Overview}

\begin{figure}[t]
\centering
\includegraphics[width=0.4\textwidth]{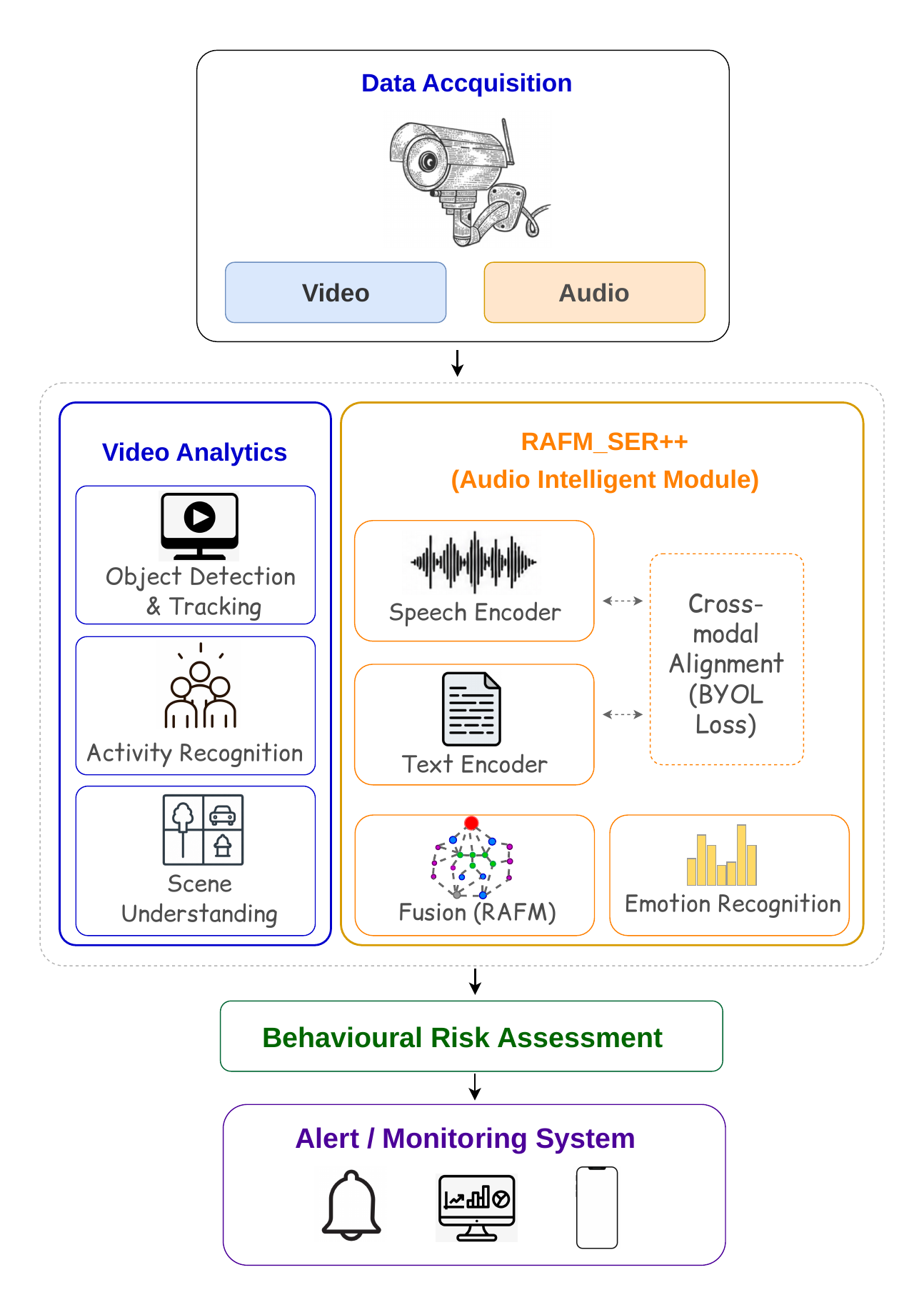}
\caption{Integration of RAFM SER++ into a multimodal surveillance pipeline.}
\label{fig:integrate}
\end{figure}

As illustrated in Fig.~\ref{fig:integrate}, RAFM\_SER++ is designed as an auxiliary audio intelligence module within a multimodal surveillance pipeline, complementing video analytics with affective speech cues for downstream behavioral monitoring.

\begin{figure*}[t]
\centering
\includegraphics[width=1.0\textwidth]{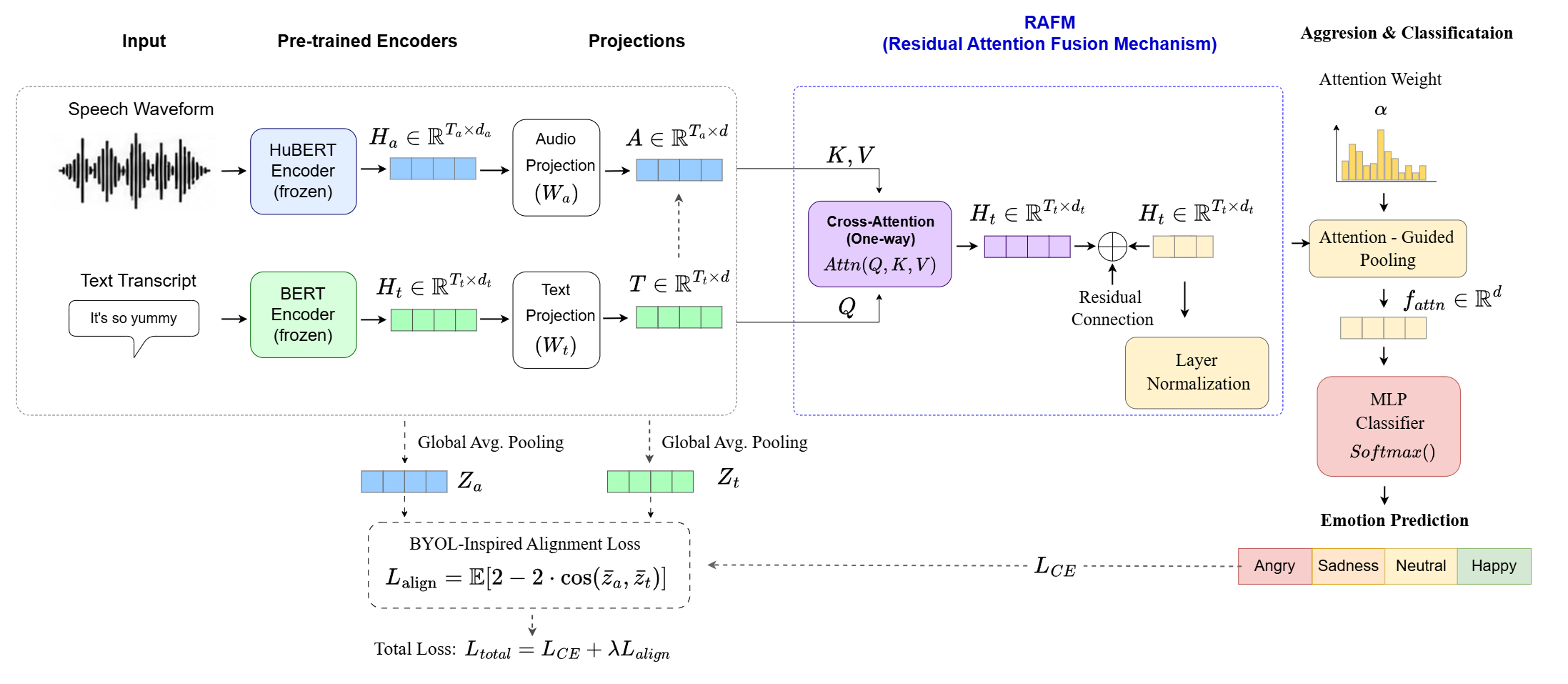}
\caption{Overview of the RAFM\_SER++ architecture, where text and speech features are fused via the proposed Residual Attention Fusion Mechanism.}
\label{fig:rafm_architecture}
\end{figure*}

RAFM\_SER++ is a lightweight multimodal SER framework designed for real-time behavioral monitoring. As shown in Fig.~\ref{fig:rafm_architecture}, the framework consists of four stages:
(i) modality-specific encoding,
(ii) projection into a shared latent space,
(iii) Residual Attention Fusion Mechanism (RAFM),
and (iv) attention-guided prediction with joint optimization.

Speech is encoded using HuBERT, while transcripts are encoded using BERT-base. Both representations are projected into a shared embedding space before fusion.

Unlike bidirectional cross-modal transformers, RAFM performs asymmetric attention where text serves as queries and speech provides keys and values, allowing affective speech cues to refine semantic representations through a residual pathway. The fused features are aggregated by attention-guided pooling and optimized jointly using cross-entropy and a BYOL-inspired alignment objective.

\subsection{Modality-Specific Encoders}

Speech is encoded using a pretrained HuBERT encoder, while transcripts are represented using BERT-base. Let
$\mathbf A$
and
$\mathbf T$
denote the acoustic and textual embeddings, respectively. Both modalities are projected into a shared latent space,
\[
\hat{\mathbf A}=AW_a+b_a,
\qquad
\hat{\mathbf T}=TW_t+b_t,
\]
to align feature dimensions before multimodal fusion.

\subsection{Residual Attention Fusion Mechanism}

Unlike conventional bidirectional transformers, RAFM performs a single asymmetric cross-attention operation where textual embeddings act as queries and acoustic embeddings serve as keys and values,
\[
Q=\hat T W_Q,\;
K=\hat A W_K,\;
V=\hat A W_V.
\]
The fused representation is computed as
\[
F=\text{LayerNorm}
(\hat T+\text{Attn}(Q,K,V)).
\]
This residual formulation allows affective speech cues to refine semantic representations while avoiding redundant bidirectional interaction, reducing computation and inference latency.

\subsection{Attention-Guided Pooling and Classification}

Speech emotion is typically conveyed through a limited number of emotionally salient words or acoustic segments rather than being uniformly distributed across an utterance. Consequently, directly applying mean or max pooling may dilute discriminative emotional cues by assigning identical importance to all temporal representations.

To address this issue, RAFM\_SER++ employs an attention-guided pooling mechanism that adaptively estimates the contribution of each fused multimodal representation. Given the fused sequence
\[
\mathbf{F}
=
\{F_1,F_2,\ldots,F_L\},
\]
where $L$ denotes the sequence length after multimodal fusion, an attention score is first computed for every token
\[
e_i
=
W_2
\tanh
(
W_1F_i+b_1
)
+b_2.
\]
The attention weights are then normalized using the Softmax function
\[
\alpha_i
=
\frac{\exp(e_i)}
{\sum_{j=1}^{L}\exp(e_j)},
\qquad
\sum_{i=1}^{L}\alpha_i=1.
\]
The utterance-level representation is obtained by weighted feature aggregation
\[
f_{\mathrm{attn}}
=
\sum_{i=1}^{L}
\alpha_iF_i.
\]
Unlike static pooling strategies, the proposed attention mechanism dynamically emphasizes emotionally informative speech--text representations while suppressing redundant contextual information. This adaptive aggregation enables the classifier to focus on localized affective evidence, which is particularly beneficial for spontaneous emotional speech where discriminative cues often appear only within short temporal regions.

Finally, the aggregated representation is passed to a lightweight multilayer perceptron $\hat y
=
\mathrm{MLP}
(f_{\mathrm{attn}})$ to predict the emotion category.

\subsection{Cross-Modal Alignment}

To improve representation consistency, RAFM\_SER++ employs a BYOL-inspired alignment loss that maximizes the cosine similarity between paired speech and text embeddings without requiring negative samples. The overall objective is
\[
L=L_{CE}+\lambda L_{align}.
\]
The alignment loss acts as a lightweight regularizer with negligible computational overhead.

\subsection{Deployment}

RAFM\_SER++ is designed as an auxiliary audio perception module for multimodal surveillance systems. The predicted emotional states provide contextual affective cues that complement visual perception modules for higher-level behavioral analysis. Rather than inferring security threats directly, the framework serves as an additional source of information for multimodal decision making while remaining suitable for resource-constrained edge deployment.

\section{EXPERIMENTS}

\subsection{Datasets}

We evaluate RAFM\_SER++ on two widely used Speech Emotion Recognition (SER) benchmarks: \textbf{IEMOCAP}~\cite{iemocap} and the \textbf{Emotional Speech Dataset (ESD)}~\cite{esd}, which provide complementary evaluation scenarios covering spontaneous conversational speech, multilingual recordings, and balanced emotional distributions.

\textbf{IEMOCAP} is a multimodal emotional dialogue corpus containing scripted and improvised conversations with synchronized speech and transcripts. Following the standard four-class protocol, we evaluate four emotion categories: \textit{anger}, \textit{sadness}, \textit{happiness} (happy + excited), and \textit{neutral}. Due to its spontaneous speech and class imbalance, IEMOCAP provides a challenging benchmark for multimodal SER.

\textbf{ESD} contains over 29 hours of bilingual (English and Mandarin) emotional speech collected from 20 speakers. Compared with IEMOCAP, ESD provides cleaner recordings and a more balanced class distribution, enabling evaluation of multilingual generalization and deployment robustness.

Although both datasets are collected under controlled recording conditions, they remain standard public benchmarks for SER and provide a consistent protocol for evaluating surveillance-oriented affective perception models.

\subsection{Implementation Details}

RAFM\_SER++ is implemented in PyTorch and optimized using Adam with $\beta_1=0.9$, $\beta_2=0.999$, $\epsilon=10^{-8}$, an initial learning rate of $10^{-4}$, and weight decay of $10^{-6}$. Models are trained for 100 epochs with a batch size of 32, and the learning rate is decayed by 0.1 every 100 epochs.

Speech and text representations are extracted using pretrained HuBERT and BERT-base encoders, respectively, then projected into a shared latent space before fusion. The final utterance-level representation is compressed into a 128-dimensional embedding and classified by a lightweight MLP.

All experiments are conducted on a single NVIDIA RTX 4090 GPU. In addition to recognition metrics, we report inference throughput in iterations per second to assess deployment efficiency.

\subsection{Evaluation Metrics}

We evaluate RAFM\_SER++ using Accuracy (ACC), Balanced Accuracy (BACC), Macro-F1, Weighted-F1, and inference throughput. Since SER datasets often exhibit class imbalance, BACC is used as the primary metric because it measures the average recall across emotion classes:

\begin{equation}
\mathrm{BACC}
=
\frac{1}{C}
\sum_{c=1}^{C}
\frac{TP_c}
{TP_c+FN_c},
\end{equation}

where $C$ is the number of classes, and $TP_c$ and $FN_c$ denote true positives and false negatives for class $c$. Macro-F1 evaluates class-balanced performance, Weighted-F1 accounts for class frequency, and throughput (\textit{it/s}) measures deployment efficiency.
\subsection{Results on ESD}

Table~\ref{tab:esd} summarizes the experimental results on the ESD dataset. RAFM\_SER++ achieves the best performance across all evaluation metrics, obtaining \textbf{95.39\%} BACC, ACC, Macro-F1, and Weighted-F1. Despite using only \textbf{3.6M} trainable parameters, the proposed framework outperforms MemoCMT (8.9M parameters) while achieving substantially faster inference (\textbf{79.60} vs. 45.19 it/s), demonstrating a superior accuracy--efficiency trade-off.

The improvement is particularly notable because it is achieved with more than a 60\% reduction in trainable parameters. These results indicate that the proposed asymmetric Residual Attention Fusion Mechanism effectively preserves complementary speech--text information without relying on computationally expensive bidirectional transformer interactions, making RAFM\_SER++ well suited for efficient multimodal SER.

\begin{table*}[!t]
\centering
\caption{Comparison of model performance with different pooling strategies on the ESD dataset.}
\label{tab:esd}
\footnotesize
\setlength{\tabcolsep}{6pt}
\renewcommand{\arraystretch}{1.12}

\begin{tabular}{llcccccc}
\toprule
\textbf{Model} &
\textbf{Pooling} &
\textbf{Params} &
\textbf{BACC (\%)} &
\textbf{ACC (\%)} &
\textbf{Macro-F1 (\%)} &
\textbf{Weighted-F1 (\%)} &
\textbf{Speed (it/s)} \\
\midrule
HuBERT-Base (S3PRL) & Mean & 0.2M & 95.10 & 95.11 & 95.10 & 95.11 & -- \\
MemoCMT & Min & 8.9M & 94.06 & 94.11 & 94.07 & 94.09 & 45.19 \\
\midrule
\textbf{RAFM-SER (BYOL)} &
\textbf{Attn-guided} &
\textbf{3.6M} &
\textbf{95.39} &
\textbf{95.39} &
\textbf{95.39} &
\textbf{95.39} &
\textbf{79.60} \\
\bottomrule
\end{tabular}
\end{table*}

\begin{figure}[!t]
\centering
\begin{minipage}[b]{0.35\textwidth}
    \centering
    \includegraphics[width=\linewidth]{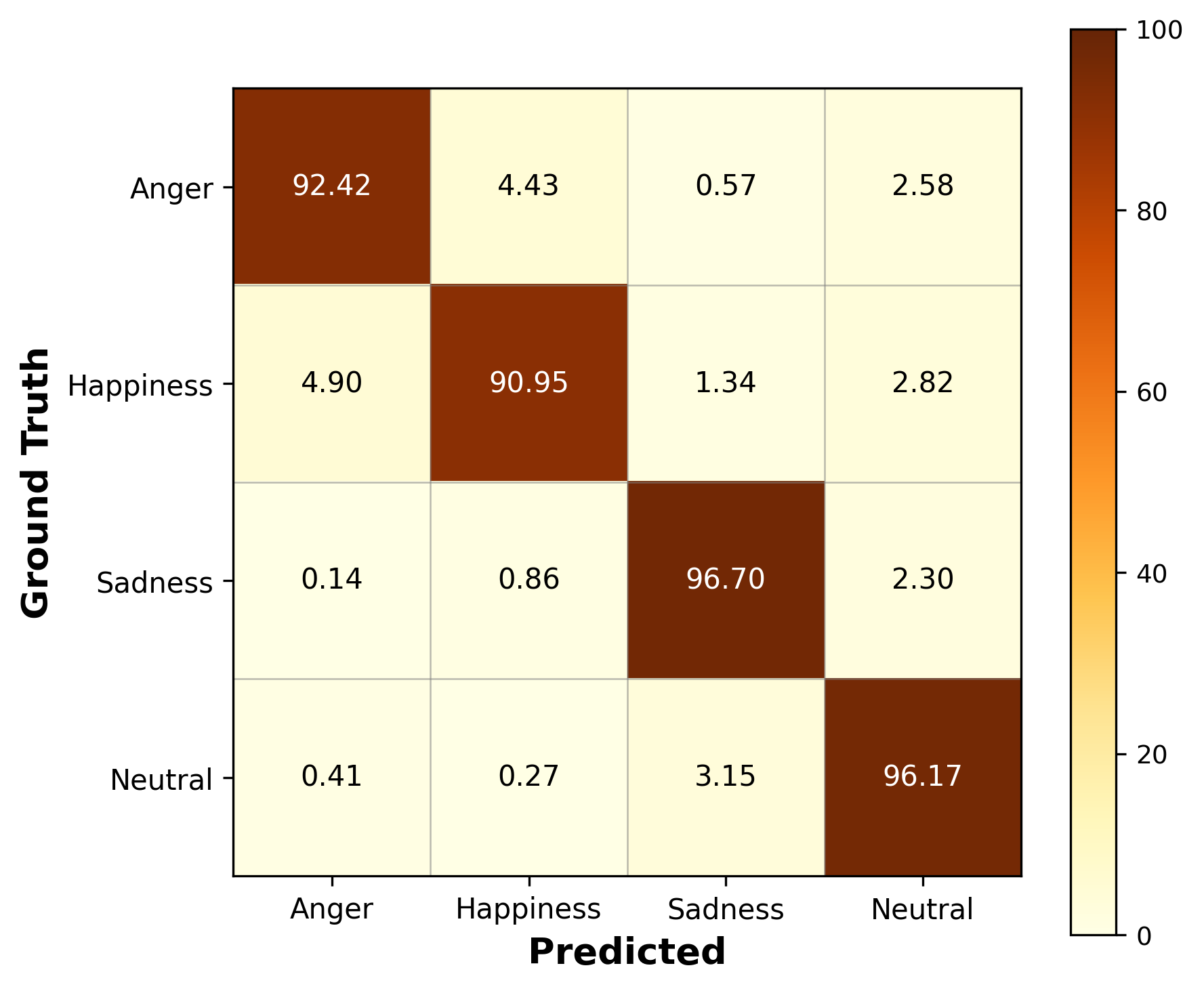}
    \caption{Confusion matrix of MemoCMT on ESD using Min pooling.}
    \label{fig:memocmtiemocap}
\end{minipage}
\hfill
\begin{minipage}[b]{0.35\textwidth}
    \centering
    \includegraphics[width=\linewidth]{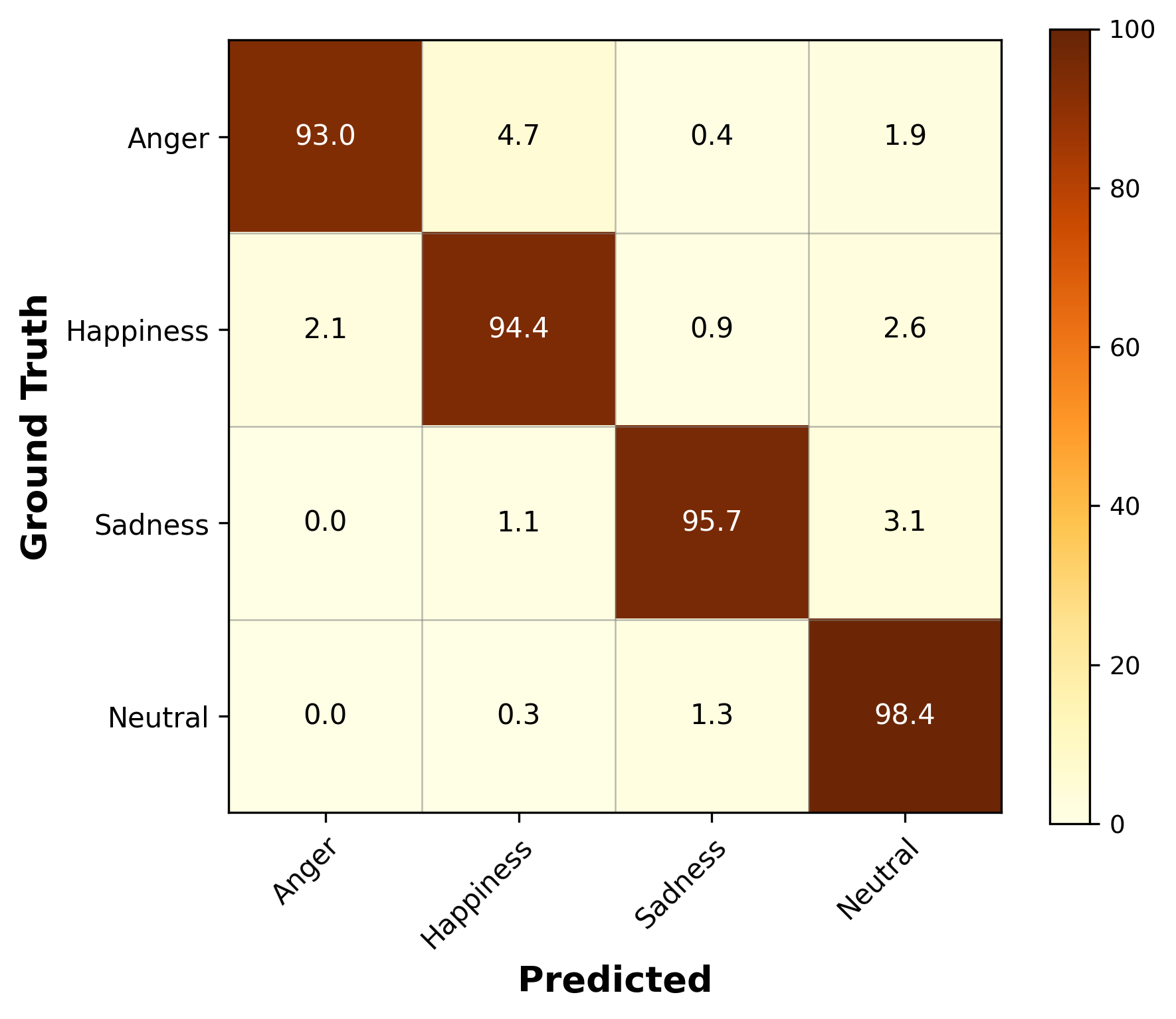}
    \caption{Confusion matrix of RAFM\_SER++ on ESD using attention-guided pooling.}
    \label{fig:rafmiemocap}
\end{minipage}
\end{figure}

Figures~\ref{fig:memocmtiemocap} and~\ref{fig:rafmiemocap} further confirm the quantitative results. Compared with MemoCMT, RAFM\_SER++ produces a cleaner confusion matrix with fewer misclassifications, particularly between the \textit{Happiness} and \textit{Neutral} classes, indicating more effective cross-modal feature integration.

\subsection{Results on IEMOCAP}

Table~\ref{tab:iemocap} reports the results on the more challenging IEMOCAP benchmark, which contains spontaneous conversational speech with greater emotional ambiguity and class imbalance than ESD.

\begin{table*}[!t]
\centering
\caption{Comparison of model performance with different pooling strategies on the IEMOCAP dataset.}
\label{tab:iemocap}
\footnotesize
\setlength{\tabcolsep}{6pt}
\renewcommand{\arraystretch}{1.12}

\begin{tabular}{llcccccc}
\toprule
\textbf{Model} &
\textbf{Pooling} &
\textbf{Params} &
\textbf{BACC (\%)} &
\textbf{ACC (\%)} &
\textbf{Macro-F1 (\%)} &
\textbf{Weighted-F1 (\%)} &
\textbf{Speed (it/s)} \\
\midrule
HuBERT-Base (S3PRL) & Mean & 0.2M & 72.39 & 72.74 & 73.03 & 72.77 & -- \\
MemoCMT & Min & 8.9M & 77.46 & 76.35 & 77.53 & 76.45 & 71.70 \\
MemoCMT & Max & 8.9M & 75.73 & 75.45 & 76.27 & 75.59 & 71.70 \\
RAFM-SER (CE) & Min & 3.6M & 74.73 & 74.85 & 74.60 & 74.40 & 73.20 \\
RAFM-SER (BYOL) & Min & 3.6M & 77.69 & 76.53 & 77.71 & 76.56 & 72.67 \\
RAFM-SER (BYOL) & Max & 3.6M & 77.60 & 76.71 & 77.29 & 76.66 & 73.05 \\
\midrule
\textbf{RAFM-SER (BYOL)} &
\textbf{Attn-guided} &
\textbf{3.6M} &
\textbf{81.10} &
\textbf{78.88} &
\textbf{79.58} &
\textbf{78.48} &
\textbf{74.28} \\
\bottomrule
\end{tabular}
\end{table*}

RAFM\_SER++ with BYOL-based alignment and attention-guided pooling achieves the best overall performance, reaching \textbf{81.10\% BACC}, \textbf{78.88\% ACC}, \textbf{79.58\% Macro-F1}, and \textbf{78.48\% Weighted-F1}. Compared with MemoCMT (Min pooling), it improves BACC by \textbf{3.64 percentage points} while reducing trainable parameters from \textbf{8.9M} to \textbf{3.6M} and maintaining slightly higher inference throughput (74.28 vs. 71.70 it/s). These results demonstrate that the proposed asymmetric fusion strategy remains effective even under more challenging conversational conditions.

\subsection{Ablation and Efficiency Analysis}

\textbf{Component Contribution Analysis.}
Table~\ref{tab:ablation} reports the contribution of each component on the IEMOCAP dataset. Introducing the BYOL-inspired alignment improves BACC from \textbf{74.73\%} to \textbf{77.69\%}, while replacing conventional pooling with the proposed attention-guided pooling further increases BACC to \textbf{81.10\%}. These results demonstrate that both cross-modal alignment and adaptive feature aggregation contribute consistently to the overall performance.

\begin{table}[!t]
\centering
\caption{Ablation study on contrastive alignment and pooling strategies.}
\label{tab:ablation}
\footnotesize
\setlength{\tabcolsep}{7pt}
\renewcommand{\arraystretch}{1.12}

\begin{tabular}{lccc}
\toprule
\textbf{Variant} & \textbf{Alignment} & \textbf{Pooling} & \textbf{BACC (\%)} \\
\midrule
RAFM-SER (CE) & No & Min & 74.73 \\
RAFM-SER (BYOL) & Yes & Min & 77.69 \\
RAFM-SER (BYOL) & Yes & Max & 77.60 \\
\midrule
\textbf{RAFM-SER (BYOL)} & \textbf{Yes} & \textbf{Attn-guided} & \textbf{81.10} \\
\bottomrule
\end{tabular}
\end{table}

\textbf{Efficiency.}
RAFM\_SER++ reduces trainable parameters by more than 60\% compared with MemoCMT while achieving higher BACC and faster inference, demonstrating its suitability for real-time deployment.

\subsection{Discussion}

The experimental results demonstrate that \textbf{RAFM\_SER++} achieves a favorable balance between recognition accuracy and deployment efficiency across both IEMOCAP and ESD. Compared with MemoCMT, the proposed framework delivers competitive or superior performance while reducing trainable parameters by more than 60\% and providing faster inference. These results indicate that effective multimodal SER does not necessarily require computationally intensive bidirectional cross-modal interaction.

A key observation is that speech and text play complementary rather than symmetric roles. RAFM\_SER++ models speech as an \emph{affective residual correction} over semantic text representations through a lightweight asymmetric attention pathway, while the BYOL-inspired alignment objective and attention-guided pooling further improve representation consistency and discriminative feature aggregation.

For surveillance applications, RAFM\_SER++ is intended as an auxiliary audio intelligence module that complements visual perception rather than replacing it. Although current experiments are conducted on benchmark SER datasets, real surveillance environments involve background noise, reverberation, multilingual speech, speaker overlap, domain shift, and ASR errors. Therefore, predicted emotions should be interpreted as contextual behavioral cues rather than direct indicators of security threats. Future work will evaluate RAFM\_SER++ under realistic audio--visual surveillance conditions and investigate adaptive modality reliability for more robust deployment.

\section{CONCLUSION}

This paper presented \textbf{RAFM\_SER++}, a lightweight deployment-oriented multimodal Speech Emotion Recognition framework for audio-visual surveillance. By combining an asymmetric Residual Attention Fusion Mechanism (RAFM), BYOL-inspired cross-modal alignment, and attention-guided pooling, it achieves effective multimodal representation learning with low computational overhead. Experimental results on the IEMOCAP and ESD benchmarks show that RAFM\_SER++ consistently achieves competitive or superior recognition performance while reducing trainable parameters by more than 60\% and providing faster inference than stronger transformer-based baselines. These findings demonstrate that lightweight asymmetric multimodal interaction effectively balances between recognition accuracy and deployment efficiency for real-time surveillance applications.

Although privacy preservation is beyond the scope of this work, the lightweight design of RAFM\_SER++ also makes it attractive for privacy-aware surveillance systems by enabling efficient on-device affective perception with reduced reliance on transmitting raw speech data. 

Future work will focus on evaluating RAFM\_SER++ under realistic surveillance conditions, including noisy environments, multilingual speech, automatic speech recognition errors, tighter audio--visual integration, and lightweight optimization for the continuous edge deployment.

\bibliographystyle{IEEEtran}
\bibliography{Bib_references}

\end{document}